# Efficiency versus Convergence of Boolean Kernels for On-Line Learning Algorithms

**Roni Khardon**                                             RONI@CS.TUFTS.EDU
*Department of Computer Science, Tufts University*
*Medford, MA 02155*

**Dan Roth**                                                  DANR@CS.UIUC.EDU
*Department of Computer Science, University of Illinois*
*Urbana, IL 61801 USA*

**Rocco A. Servedio**                                        ROCCO@CS.COLUMBIA.EDU
*Department of Computer Science, Columbia University*
*New York, NY 10025*

## Abstract

The paper studies machine learning problems where each example is described using a set of Boolean features and where hypotheses are represented by linear threshold elements. One method of increasing the expressiveness of learned hypotheses in this context is to expand the feature set to include conjunctions of basic features. This can be done explicitly or where possible by using a kernel function. Focusing on the well known Perceptron and Winnow algorithms, the paper demonstrates a tradeoff between the computational efficiency with which the algorithm can be run over the expanded feature space and the generalization ability of the corresponding learning algorithm.

We first describe several kernel functions which capture either limited forms of conjunctions or all conjunctions. We show that these kernels can be used to efficiently run the Perceptron algorithm over a feature space of exponentially many conjunctions; however we also show that using such kernels, the Perceptron algorithm can provably make an exponential number of mistakes even when learning simple functions.

We then consider the question of whether kernel functions can analogously be used to run the multiplicative-update Winnow algorithm over an expanded feature space of exponentially many conjunctions. Known upper bounds imply that the Winnow algorithm can learn Disjunctive Normal Form (DNF) formulae with a polynomial mistake bound in this setting. However, we prove that it is computationally hard to simulate Winnow's behavior for learning DNF over such a feature set. This implies that the kernel functions which correspond to running Winnow for this problem are not efficiently computable, and that there is no general construction that can run Winnow with kernels.

## 1. Introduction

The problem of classifying objects into one of two classes being "positive" and "negative" examples of a concept is often studied in machine learning. The task in machine learning is to extract such a classifier from given pre-classified examples - the problem of learning from data. When each example is represented by a set of $n$ numerical features, an example can be seen as a point in Euclidean space $\Re^n$. A common representation for classifiers in this case is a hyperplane of dimension $(n-1)$ which splits the domain of examples into two areas of positive and negative examples. Such a representation is known as a *linear threshold function*, and many learning algorithms that output a hypothesis represented in this manner have been developed, analyzed, implemented, and applied in practice. Of particular interest in this paper are the well known Perceptron (Rosenblatt, 1958; Block, 1962; Novikoff, 1963) and Winnow (Littlestone, 1988) algorithms that have been intensively studied in the literature.

It is also well known that the expressiveness of linear threshold functions is quite limited (Minsky & Papert, 1968). Despite this fact, both Perceptron and Winnow have been applied successfully in recent years to several large scale real world classification problems. As one example, the SNoW system (Roth, 1998; Carlson, Cumby, Rosen, & Roth, 1999) has successfully applied variations of Perceptron and Winnow to problems in natural language processing. The SNoW system extracts basic Boolean features $x_1, \ldots, x_n$ from labeled pieces of text data in order to represent the examples, thus the features have numerical values restricted to $\{0, 1\}$. There are several ways to enhance the set of basic features $x_1, \ldots, x_n$ for Perceptron or Winnow. One idea is to expand the set of basic features $x_1, \ldots, x_n$ using conjunctions such as $(x_1 \wedge \overline{x}_3 \wedge x_4)$ and use these expanded higher-dimensional examples, in which each conjunction plays the role of a basic feature, as the examples for Perceptron or Winnow. This is in fact the approach which the SNoW system takes running Perceptron or Winnow over a space of restricted conjunctions of these basic features. This idea is closely related to the use of kernel methods, see e.g. the book of Cristianini and Shawe-Taylor (2000), where a feature expansion is done implicitly through the kernel function. The approach clearly leads to an increase in expressiveness and thus may improve performance. However, it also dramatically increases the number of features (from $n$ to $3^n$ if all conjunctions are used), and thus may adversely affect both the computation time and convergence rate of learning. The paper provides a theoretical study of the performance of Perceptron and Winnow when run over expanded feature spaces such as these.

## 1.1 Background: On-Line Learning with Perceptron and Winnow

Before describing our results, we recall some necessary background on the on-line learning model (Littlestone, 1988) and the Perceptron and Winnow algorithms.

Given an instance space $X$ of possible examples, a concept is a mapping of instances into one of two (or more) classes. A concept class $C \subseteq 2^X$ is simply a set of concepts. In on-line learning a concept class $C$ is fixed in advance and an adversary can pick a concept $c \in C$. The learning is then modeled as a repeated game where in each iteration the adversary picks an example $x \in X$, the learner gives a guess for the value of $c(x)$ and is then told the correct value. We count one mistake for each iteration in which the value is not predicted correctly. A learning algorithm learns a concept class $C$ with mistake bound $M$ if for any choice of $c \in C$ and any (arbitrarily long) sequence of examples, the learner is guaranteed to make at most $M$ mistakes.

In this paper we consider the case where the examples are given by Boolean features, that is $X = \{0, 1\}^n$, and we have two class labels denoted by $-1$ and $1$. Thus for $x \in \{0, 1\}^n$, a labeled example $\langle x, 1 \rangle$ is a positive example, and a labeled example $\langle x, -1 \rangle$ is a negative example. The concepts we consider are built using logical combinations of the $n$ base features and we are interested in mistake bounds that are polynomial in $n$.

### 1.1.1 PERCEPTRON

Throughout its execution Perceptron maintains a weight vector $w \in \Re^N$ which is initially $(0, \ldots, 0)$. Upon receiving an example $x \in \Re^N$ the algorithm predicts according to the linear threshold function $w \cdot x \geq 0$. If the prediction is 1 and the label is $-1$ (false positive prediction) then the vector $w$ is set to $w - x$, while if the prediction is $-1$ and the label is 1 (false negative) then $w$ is set to $w + x$. No change is made to $w$ if the prediction is correct. Many variants of this basic algorithm have been proposed and studied and in particular one can add a non zero threshold as well as a learning rate that controls the size of update to $w$. Some of these are discussed further in Section 3.

The famous Perceptron Convergence Theorem (Rosenblatt, 1958; Block, 1962; Novikoff, 1963) bounds the number of mistakes which the Perceptron algorithm can make:

**Theorem 1** *Let* $\langle x^1, y_1 \rangle, \ldots, \langle x^t, y_t \rangle$ *be a sequence of labeled examples with* $x^i \in \Re^N$, $\|x^i\| \leq R$ *and* $y_i \in \{-1, 1\}$ *for all* $i$. *Let* $u \in \Re^N, \xi > 0$ *be such that* $y_i(u \cdot x^i) \geq \xi$ *for all* $i$. *Then Perceptron makes at most* $\frac{R^2\|u\|^2}{\xi^2}$ *mistakes on this example sequence.*

### 1.1.2 WINNOW

The Winnow algorithm (Littlestone, 1988) has a very similar structure. Winnow maintains a hypothesis vector $w \in \Re^N$ which is initially $w = (1, \ldots, 1)$. Winnow is parameterized by a promotion factor $\alpha > 1$ and a threshold $\theta > 0$; upon receiving an example $x \in \{0, 1\}^N$ Winnow predicts according to the threshold function $w \cdot x \geq \theta$. If the prediction is 1 and the label is $-1$ then for all $i$ such that $x_i = 1$ the value of $w_i$ is set to $w_i/\alpha$; this is a *demotion* step. If the prediction is $-1$ and the label is 1 then for all $i$ such that $x_i = 1$ the value of $w_i$ is set to $\alpha w_i$; this is a *promotion* step. No change is made to $w$ if the prediction is correct.

For our purposes the following mistake bound, implicit in Littlestone's work (1988), is of interest:

**Theorem 2** *Let the target function be a $k$-literal monotone disjunction* $f(x_1, \ldots, x_N) = x_{i_1} \vee \cdots \vee x_{i_k}$. *For any sequence of examples in* $\{0, 1\}^N$ *labeled according to* $f$ *the number of prediction mistakes made by Winnow$(\alpha, \theta)$ is at most* $\frac{\alpha}{\alpha-1} \cdot \frac{N}{\theta} + k(\alpha+1)(1 + \log_\alpha \theta)$.

## 1.2 Our Results

We are interested in the computational efficiency and convergence of the Perceptron and Winnow algorithms when run over expanded feature spaces of conjunctions. Specifically, we study the use of kernel functions to expand the feature space and thus enhance the learning abilities of Perceptron and Winnow; we refer to these enhanced algorithms as *kernel Perceptron* and *kernel Winnow*.

Our first result (cf. also the papers of Sadohara, 1991; Watkins, 1999; and Kowalczyk et al., 2001) uses kernel functions to show that it is possible to efficiently run the kernel Perceptron algorithm over an exponential number of conjunctive features.

**Result 1:** (see Theorem 3) There is an algorithm that simulates Perceptron over the $3^n$-dimensional feature space of all conjunctions of $n$ basic features. Given a sequence of $t$ labeled examples in $\{0,1\}^n$ the prediction and update for each example take poly$(n, t)$ time steps. We also prove variants of this result in which the expanded feature space consists of all monotone conjunctions or all conjunctions of some bounded size.

This result is closely related to one of the main open problems in learning theory: efficient learnability of disjunctions of conjunctions, or DNF (Disjunctive Normal Form) expressions.[1] Since linear threshold elements can represent disjunctions (e.g. $x_1 \vee x_2 \vee x_3$ is true iff $x_1 + x_2 + x_3 \geq 1$), Theorem 1 and Result 1 imply that kernel Perceptron can be used to learn DNF. However, in this framework the values of $N$ and $R$ in Theorem 1 can be exponentially large (note that we have $N = 3^n$ and $R = 2^{n/2}$ if all conjunctions are used), and hence the mistake bound given by Theorem 1 is exponential rather than polynomial in $n$. The question thus arises whether the exponential upper bound implied by Theorem 1 is essentially tight for the kernel Perceptron algorithm in the context of DNF learning. We give an affirmative answer, thus showing that kernel Perceptron cannot efficiently learn DNF.

**Result 2:** There is a monotone DNF $f$ over $x_1, \ldots, x_n$ and a sequence of examples labeled according to $f$ which causes the kernel Perceptron algorithm to make $2^{\Omega(n)}$ mistakes. This result holds for generalized versions of the Perceptron algorithm where a fixed or updated threshold and a learning rate are used. We also give a variant of this result showing that kernel Perceptron fails in the Probably Approximately Correct (PAC) learning model (Valiant, 1984) as well.

Turning to Winnow, an attractive feature of Theorem 2 is that for suitable $\alpha, \theta$ the bound is logarithmic in the total number of features $N$ (e.g. $\alpha = 2$ and $\theta = N$). Therefore, as noted by several researchers (Maass & Warmuth, 1998), if a Winnow analogue of Theorem 3 could be obtained this would imply that DNF can be learned by a computationally efficient algorithm with a poly$(n)$-mistake bound. However, we give strong evidence that no such Winnow analogue of Theorem 3 can exist.

**Result 3:** There is no polynomial time algorithm which simulates Winnow over exponentially many monotone conjunctive features for learning monotone DNF unless every problem in the complexity class #P can be solved in polynomial time. This result holds for a wide range of parameter settings in the Winnow algorithm.

We observe that, in contrast to this negative result, Maass and Warmuth have shown that the Winnow algorithm can be simulated efficiently over exponentially many conjunctive features for learning some simple geometric concept classes (Maass & Warmuth, 1998).

Our results thus indicate a tradeoff between computational efficiency and convergence of kernel algorithms for rich classes of Boolean functions such as DNF formulas; the kernel

---

1. Angluin (1990) proved that DNF expressions cannot be learned efficiently using equivalence queries whose hypotheses are themselves DNF expressions. Since the model of exact learning from equivalence queries only is equivalent to the mistake bound model which we consider in this paper, her result implies that no online algorithm which uses DNF formulas as hypotheses can efficiently learn DNF. However, this result does not preclude the efficient learnability of DNF using a different class of hypotheses. The kernel Perceptron algorithm generates hypotheses which are thresholds of conjunctions rather than DNF formulas, and thus Angluin's negative results do not apply here.

Perceptron algorithm is computationally efficient to run but has exponentially slow convergence, whereas kernel Winnow has rapid convergence but seems to require exponential runtime.

## 2. Kernel Perceptron with Many Features

It is well known that the hypothesis $w$ of the Perceptron algorithm is a linear combination of the previous examples on which mistakes were made (Cristianini & Shaw-Taylor, 2000). More precisely, if we let $L(v) \in \{-1, 1\}$ denote the label of example $v$, then we have that $w = \sum_{v \in M} L(v) v$ where $M$ is the set of examples on which the algorithm made a mistake. Thus the prediction of Perceptron on $x$ is 1 iff $w \cdot x = (\sum_{v \in M} L(v) v) \cdot x = \sum_{v \in M} L(v)(v \cdot x) \geq 0$.

For an example $x \in \{0, 1\}^n$ let $\phi(x)$ denote its transformation into an enhanced feature space such as the space of all conjunctions. To run the Perceptron algorithm over the enhanced space we must predict 1 iff $w^\phi \cdot \phi(x) \geq 0$ where $w^\phi$ is the weight vector in the enhanced space; from the above discussion this holds iff $\sum_{v \in M} L(v)(\phi(v) \cdot \phi(x)) \geq 0$. Denoting $K(v, x) = \phi(v) \cdot \phi(x)$ this holds iff $\sum_{v \in M} L(v) K(v, x) \geq 0$.

Thus we never need to construct the enhanced feature space explicitly; in order to run Perceptron we need only be able to compute the kernel function $K(v, x)$ efficiently. This is the idea behind all so-called kernel methods, which can be applied to any algorithm (such as support vector machines) whose prediction is a function of inner products of examples. A more detailed discussion is given in the book of Cristianini and Shawe-Taylor (2000). Thus the next theorem is simply obtained by presenting a kernel function capturing all conjunctions.

**Theorem 3** *There is an algorithm that simulates Perceptron over the feature spaces of (1) all conjunctions, (2) all monotone conjunctions, (3) conjunctions of size $\leq k$, and (4) monotone conjunctions of size $\leq k$. Given a sequence of $t$ labeled examples in $\{0, 1\}^n$ the prediction and update for each example take poly$(n, t)$ time steps.*

**Proof:** For case (1) $\phi(\cdot)$ includes all $3^n$ conjunctions (with positive and negative literals) and $K(x, y)$ must compute the number of conjunctions which are true in both $x$ and $y$. Clearly, any literal in such a conjunction must satisfy both $x$ and $y$ and thus the corresponding bit in $x, y$ must have the same value. Thus each conjunction true in both $x$ and $y$ corresponds to a subset of such bits. Counting all these conjunctions gives $K(x, y) = 2^{\text{same}(x,y)}$ where same(x, y) is the number of original features that have the same value in $x$ and $y$, i.e. the number of bit positions $i$ which have $x_i = y_i$. This kernel has been obtained independently by Sadohara (2001).

To express all monotone monomials as in (2) we take $K(x, y) = 2^{|x \cap y|}$ where $|x \cap y|$ is the number of active features common to both $x$ and $y$, i.e. the number of bit positions which have $x_i = y_i = 1$.

Similarly, for case (3) the number of conjunctions that satisfy both $x$ and $y$ is $K(x, y) = \sum_{l=0}^{k} \binom{\text{same}(x,y)}{1}$. This kernel is reported also by Watkins (1999). For case (4) we have $K(x, y) = \sum_{l=0}^{k} \binom{|x \cap y|}{l}$. □

## 3. Kernel Perceptron with Many Mistakes

In this section we describe a simple monotone DNF target function and a sequence of labeled examples which causes the monotone monomials kernel Perceptron algorithm to make exponentially many mistakes.

For $x, y \in \{0, 1\}^n$ we write $|x|$ to denote the number of 1's in $x$ and, as described above, $|x \cap y|$ to denote the number of bit positions $i$ which have $x_i = y_i = 1$. We need the following well-known tail bound on sums of independent random variables which can be found in, e.g., Section 9.3 of the book by Kearns and Vazirani (1994):

**Fact 4** *Let $X_1, \ldots, X_m$ be a sequence of $m$ independent 0/1-valued random variables, each of which has $E[X_i] = p$. Let $X$ denote $\sum_{i=1}^{m} X_i$, so $E[X] = pm$. Then for $0 \leq \gamma \leq 1$, we have*

$$\Pr[X > (1 + \gamma)pm] \leq e^{-mp\gamma^2/3} \qquad and \qquad \Pr[X < (1 - \gamma)pm] \leq e^{-mp\gamma^2/2}.$$

We also use the following combinatorial property:

**Lemma 5** *There is a set $S$ of $n$-bit strings $S = \{x^1, \ldots, x^t\} \subset \{0, 1\}^n$ with $t = e^{n/9600}$ such that $|x^i| = n/20$ for $1 \leq i \leq t$ and $|x^i \cap x^j| \leq n/80$ for $1 \leq i < j \leq t$.*

**Proof:** We use the probabilistic method. For each $i = 1, \ldots, t$ let $x^i \in \{0, 1\}^n$ be chosen by independently setting each bit to 1 with probability $1/10$. For any $i$ it is clear that $E[|x^i|] = n/10$. Applying Fact 4, we have that $\Pr[|x^i| < n/20] \leq e^{-n/80}$, and thus the probability that any $x^i$ satisfies $|x^i| < n/20$ is at most $te^{-n/80}$. Similarly, for any $i \neq j$ we have $E[|x^i \cap x^j|] = n/100$. Applying Fact 4 we have that $\Pr[|x^i \cap x^j| > n/80] \leq e^{-n/4800}$, and thus the probability that any $x^i, x^j$ with $i \neq j$ satisfies $|x^i \cap x^j| > n/80$ is at most $\binom{t}{2} e^{-n/4800}$. For $t = e^{n/9600}$ the value of $\binom{t}{2} e^{-n/4800} + te^{-n/80}$ is less than 1. Thus for some choice of $x^1, \ldots, x^t$ we have each $|x^i| \geq n/20$ and $|x^i \cap x^j| \leq n/80$. For any $x^i$ which has $|x^i| > n/20$ we can set $|x^i| - n/20$ of the 1s to 0s, and the lemma is proved. $\qquad \square$

Now using the previous lemma we can construct a difficult data set for kernel Perceptron:

**Theorem 6** *There is a monotone DNF $f$ over $x_1, \ldots, x_n$ and a sequence of examples labeled according to $f$ which causes the kernel Perceptron algorithm to make $2^{\Omega(n)}$ mistakes.*

**Proof:** The target DNF with which we will use is very simple: it is the single conjunction $x_1 x_2 \ldots x_n$. While the original Perceptron algorithm over the $n$ features $x_1, \ldots, x_n$ is easily seen to make at most $\text{poly}(n)$ mistakes for this target function, we now show that the monotone kernel Perceptron algorithm which runs over a feature space of all $2^n$ monotone monomials can make $2 + e^{n/9600}$ mistakes.

Recall that at the beginning of the Perceptron algorithm's execution all $2^n$ coordinates of $w^\phi$ are 0. The first example is the negative example $0^n$. The only monomial true in this example is the empty monomial which is true in every example. Since $w^\phi \cdot \phi(x) = 0$ Perceptron incorrectly predicts 1 on this example. The resulting update causes the coefficient $w_\emptyset^\phi$ corresponding to the empty monomial to become $-1$ but all $2^n - 1$ other coordinates of $w^\phi$ remain 0. The next example is the positive example $1^n$. For this example we have $w^\phi \cdot \phi(x) = -1$ so Perceptron incorrectly predicts $-1$. Since all $2^n$ monotone conjunctions are satisfied by this example the resulting update causes $w_\emptyset^\phi$ to become 0 and all $2^n - 1$ other coordinates of $w^\phi$ to become 1. The next $e^{n/9600}$ examples are the vectors $x^1, \ldots, x^t$ described in Lemma 5. Since each such example has $|x^i| = n/20$ each example is negative; however as we now show the Perceptron algorithm will predict 1 on each of these examples.

Fix any value $1 \le i \le e^{n/9600}$ and consider the hypothesis vector $w^\phi$ just before example $x^i$ is received. Since $|x^i| = n/20$ the value of $w^\phi \cdot \phi(x^i)$ is a sum of the $2^{n/20}$ different coordinates $w_T^\phi$ which correspond to the monomials satisfied by $x^i$. More precisely we have $w^\phi \cdot \phi(x^i) = \sum_{T \in A_i} w_T^\phi + \sum_{T \in B_i} w_T^\phi$ where $A_i$ contains the monomials which are satisfied by $x^i$ and $x^j$ for some $j \ne i$ and $B_i$ contains the monomials which are satisfied by $x^i$ but no $x^j$ with $j \ne i$. We lower bound the two sums separately.

Let $T$ be any monomial in $A_i$. By Lemma 5 any $T \in A_i$ contains at most $n/80$ variables and thus there can be at most $\sum_{r=0}^{n/80} \binom{n/20}{r}$ monomials in $A_i$. Using the well known bound $\sum_{j=0}^{\alpha \ell} \binom{\ell}{j} = 2^{(H(\alpha) + o(1))\ell}$ where $0 < \alpha \le 1/2$ and $H(p) = -p \log p - (1-p) \log(1-p)$ is the binary entropy function, which can be found e.g. as Theorem 1.4.5 of the book by Van Lint (1992), there can be at most $2^{0.8113 \cdot (n/20) + o(n)} < 2^{0.041n}$ terms in $A_i$. Moreover the value of each $w_T^\phi$ must be at least $-e^{n/9600}$ since $w_T^\phi$ decreases by at most 1 for each example, and hence $\sum_{T \in A_i} w_T^\phi \ge -e^{n/9600} 2^{0.041n} > -2^{0.042n}$. On the other hand, any $T \in B_i$ is false in all other examples and therefore $w_T^\phi$ has not been demoted and $w_T^\phi = 1$. By Lemma 5 for any $r > n/80$ every $r$-variable monomial satisfied by $x_i$ must belong to $B_i$, and hence $\sum_{T \in B_i} w_T^\phi \ge \sum_{r=n/80+1}^{n/20} \binom{n/20}{r} > 2^{0.049n}$. Combining these inequalities we have $w \cdot x^i \ge -2^{0.042n} + 2^{0.049n} > 0$ and hence the Perceptron prediction on $x^i$ is 1. $\qquad \square$

**Remark 7** At first sight it might seem that the result is limited to a simple special case of the perceptron algorithm. Several variations exist that use: an added feature with a fixed value that enables the algorithm to update the threshold indirectly (via a weight $\hat{w}$), a non zero fixed (initial) threshold $\theta$, and a learning rate $\alpha$, and in particular all these three can be used simultaneously. The generalized algorithm predicts according to the hypothesis $w \cdot x + \hat{w} \ge \theta$ and updates $w \leftarrow w + \alpha x$ and $\hat{w} \leftarrow \hat{w} + \alpha$ for promotions and similarly for demotions. We show here that exponential lower bounds on the number of mistakes can be derived for the more general algorithm as well. First, note that since our kernel includes a feature for the empty monomial which is always true, the first parameter is already accounted for. For the other two parameters note that there is a degree of freedom between the learning rate $\alpha$ and fixed threshold $\theta$ since multiplying both by the same factor does not change the hypothesis and therefore it suffices to consider the threshold only. We consider several cases for the value of the threshold. If $\theta$ satisfies $0 \le \theta \le 2^{0.047}$ then we use the same sequence of examples. After the first two examples the algorithm makes a promotion on $1^n$ (it may or may not update on $0^n$ but that is not important). For the examples in the sequence the bounds on $\sum_{T \in A_i} w_T^\phi$ and $\sum_{T \in B_i} w_T^\phi$ are still valid so the final inequality in the proof becomes $w \cdot x^i \ge -2^{0.042n} + 2^{0.049n} > 2^{0.047n}$ which is true for sufficiently large $n$. If $\theta > 2^{0.047n}$ then we can construct the following scenario. We use the function $f = x_1 \lor x_2 \lor \ldots \lor x_n$, and the sequence of examples includes $\frac{\theta}{2} - 1$ repetitions of the same example $x$ where the first bit is 1 and all other bits are 0. The example $x$ satisfies exactly 2 monomials and therefore the algorithm will make mistakes on all the examples in the sequence. If $\theta < 0$ then the initial hypothesis misclassifies $0^n$. We start the example sequence by repeating the example $0^n$ until it is classified correctly, that is $\lceil -\theta \rceil$ times. If the threshold is large in absolute value e.g. $\theta < -2^{0.042n}$ we are done. Otherwise we continue with the example $1^n$. Since all weights except for the empty monomial are zero at this stage the examples $0^n$ and $1^n$ are classified in the same way so $1^n$ is misclassified and therefore the algorithm makes a promotion. The argument for the rest of the sequence is as above (except for adding a term for the empty monomial) and the final inequality becomes $w \cdot x^i \geq -2^{0.042n} - 2^{0.042n} + 2^{0.049n} > -2^{0.042n}$ so each of the examples is misclassified. Thus in all cases kernel Perceptron may make an exponential number of mistakes.

## 3.1 A Negative Result for the PAC Model

The proof above can be adapted to give a negative result for kernel Perceptron in the PAC learning model (Valiant, 1984). In this model each example $x$ is independently drawn from a fixed probability distribution $\mathcal{D}$ and with high probability the learner must construct a hypothesis $h$ which has high accuracy relative to the target concept $c$ under distribution $\mathcal{D}$. See the Kearns-Vazirani text (1994) for a detailed discussion of the PAC learning model.

Let $\mathcal{D}$ be the probability distribution over $\{0,1\}^n$ which assigns weight $1/4$ to the example $0^n$, weight $1/4$ to the example $1^n$, and weight $\frac{1}{2}\frac{1}{e^{n/9600}}$ to each of the $e^{n/9600}$ examples $x^1, \ldots, x^t$.

**Theorem 8** *If kernel Perceptron is run using a sample of polynomial size $p(n)$ then with probability at least $1/16$ the error of its final hypothesis is at least $0.49$.*

**Proof:** With probability $1/16$, the first two examples received from $\mathcal{D}$ will be $0^n$ and then $1^n$. Thus, with probability $1/16$, after two examples (as in the proof above) the Perceptron algorithm will have $w_\emptyset^\phi = 0$ and all other coefficients of $w^\phi$ equal to 1.

Consider the sequence of examples following these two examples. First note that in any trial, any occurrence of an example other than $1^n$ (i.e. any occurrence either of some $x^i$ or of the $0^n$ example) can decrease $\sum_{T \subseteq [n]} w_T^\theta$ by at most $2^{n/20}$. Since after the first two examples we have $w^\phi \cdot \phi(1^n) = \sum_{T \subseteq [n]} w_T^\theta = 2^n - 1$, it follows that at least $2^{19n/20} - 1$ more examples must occur before the $1^n$ example will be incorrectly classified as a negative example. Since we will only consider the performance of the algorithm for $p(n) < 2^{19n/20} - 1$ steps, we may ignore all subsequent occurrences of $1^n$ since they will not change the algorithm's hypothesis.

Now observe that on the first example which is not $1^n$ the algorithm will perform a demotion resulting in $w_\emptyset^\phi = -1$ (possibly changing other coefficients as well). Since no promotions will be performed on the rest of the sample, we get $w_\emptyset^\phi \leq -1$ for the rest of the learning process. It follows that all future occurrences of the example $0^n$ are correctly classified and thus we may ignore them as well.

Considering examples $x^i$ from the sequence constructed above, we may ignore any example that is correctly classified since no update is made on it. It follows that when the perceptron algorithm has gone over all examples, its hypothesis is formed by demotions on examples in the sequence of $x^i$'s. The only difference from the scenario above is that the algorithm may make several demotions on the same example if it occurs multiple times in the sample. However, an inspection of the proof above shows that for any $x^i$ that has not been seen by the algorithm, the bounds on $\sum_{T \in A_i} w_T^\phi$ and $\sum_{T \in B_i} w_T^\phi$ are still valid and therefore $x^i$ will be misclassified. Since the sample is of size $p(n)$ and the sequence is of size $e^{n/9600}$ the probability weight of examples in the sample is at most 0.01 for sufficiently large $n$ so the error of the hypothesis is at least 0.49. □

## 4. Computational Hardness of Kernel Winnow

In this section, for $x \in \{0, 1\}^n$ we let $\phi(x)$ denote the $(2^n - 1)$-element vector whose coordinates are all nonempty monomials (monotone conjunctions) over $x_1, \ldots, x_n$. We say that a sequence of labeled examples $\langle x^1, b_1 \rangle, \ldots, \langle x^t, b_t \rangle$ is *monotone consistent* if it is consistent with some monotone function, i.e. $x_k^i \leq x_k^j$ for all $k = 1, \ldots, n$ implies $b_i \leq b_j$. If $S$ is monotone consistent and has $t$ labeled examples then clearly there is a monotone DNF formula consistent with $S$ which contains at most $t$ conjunctions. We consider the following problem:

**KERNEL WINNOW PREDICTION$(\alpha, \theta)$ (KWP)**
**Instance:** Monotone consistent sequence $S = \langle x^1, b_1 \rangle, \ldots, \langle x^t, b_t \rangle$ of labeled examples with each $x^i \in \{0, 1\}^m$ and each $b_i \in \{-1, 1\}$; unlabeled example $z \in \{0, 1\}^m$.
**Question:** Is $w^\phi \cdot \phi(z) \geq \theta$, where $w^\phi$ is the $N = (2^m - 1)$-dimensional hypothesis vector generated by running Winnow$(\alpha, \theta)$ on the example sequence $\langle \phi(x^1), b_1 \rangle, \ldots \langle \phi(x^t), b_t \rangle$?

In order to run Winnow over all $2^m - 1$ nonempty monomials to learn monotone DNF, one must be able to solve KWP efficiently. Our main result in this section is a proof that KWP is computationally hard for a wide range of parameter settings which yield a polynomial mistake bound for Winnow via Theorem 2.

Recall that $\#P$ is the class of all counting problems associated with $NP$ decision problems; it is well known that if every function in $\#P$ is computable in polynomial time then $P = NP$. See the book of Papadimitriou (1994) or the paper of Valiant (1979) for details on $\#P$. The following problem is $\#P$-hard (Valiant, 1979):

**MONOTONE 2-SAT (M2SAT)**
**Instance:** Monotone 2-CNF Boolean formula $F = c_1 \wedge c_2 \wedge \ldots \wedge c_r$ with $c_i = (y_{i_1} \vee y_{i_2})$ and each $y_{i_j} \in \{y_1, \ldots, y_n\}$; integer $K$ such that $1 \leq K \leq 2^n$.
**Question:** Is $|F^{-1}(1)| \geq K$, i.e. does $F$ have at least $K$ satisfying assignments in $\{0, 1\}^n$?

**Theorem 9** *Fix any $\epsilon > 0$. Let $N = 2^m - 1$, let $\alpha \geq 1 + 1/m^{1-\epsilon}$, and let $\theta \geq 1$ be such that $\max(\frac{\alpha}{\alpha-1} \cdot \frac{N}{\theta}, (\alpha+1)(1 + \log_\alpha \theta)) = poly(m)$. If there is a polynomial time algorithm for KWP$(\alpha, \theta)$, then every function in $\#P$ is computable in polynomial time.*

**Proof:** For $N, \alpha$ and $\theta$ as described in the theorem a routine calculation shows that

$$1 + 1/m^{1-\epsilon} \leq \alpha \leq poly(m) \quad \text{and} \quad \frac{2^m}{poly(m)} \leq \theta \leq 2^{poly(m)}. \tag{1}$$

The proof is a reduction from the problem M2SAT. The high level idea of the proof is simple: let $(F, K)$ be an instance of M2SAT where $F$ is defined over variables $y_1, \ldots, y_n$. The Winnow algorithm maintains a weight $w_T^\phi$ for each monomial $T$ over variables $x_1, \ldots, x_m$. We define a 1-1 correspondence between these monomials $T$ and truth assignments $y^T \in \{0, 1\}^n$ for $F$, and we give a sequence of examples for Winnow which causes $w_T^\phi \approx 0$ if $F(y^T) = 0$ and $w_T^\phi = 1$ if $F(y^T) = 1$. The value of $w^\phi \cdot \phi(z)$ is thus related to $|F^{-1}(1)|$. Note that if we could control $\theta$ as well this would be sufficient since we could use $\theta = K$ and the result will follow. However $\theta$ is a parameter of the algorithm. We therefore have to make additional updates so that $w^\phi \cdot \phi(z) \approx \theta + (|F^{-1}(1)| - K)$ so that $w^\phi \cdot \phi(z) \geq \theta$ if and only if $|F^{-1}(1)| \geq K$. The details are somewhat involved since we must track the resolution of approximations of the different values so that the final inner product will indeed give a correct result with respect to the threshold.

**General setup of the construction.** In more detail, let

- $U = n + 1 + \lceil (\lceil \log_\alpha 4 \rceil + 1) \log \alpha \rceil$,

- $V = \lceil \frac{n+1}{\log \alpha} \rceil + 1$,

- $W = \lceil \frac{U+2}{\log \alpha} \rceil + 1$

and let $m$ be defined as

$$m = n + U + 6Vn^2 + 6UW + 3. \tag{2}$$

Since $\alpha \geq 1 + 1/m^{1-\epsilon}$, using the fact that $\log(1 + x) \geq x/2$ for $0 < x < 1$ we have that $\log \alpha \geq 1/(2m^{1-\epsilon})$, and from this it easily follows that $m$ as specified above is polynomial in $n$. We describe a polynomial time transformation which maps an $n$-variable instance $(F, K)$ of M2SAT to an $m$-variable instance $(S, z)$ of KWP$(\alpha, \theta)$ where $S = \langle x^1, b_1 \rangle, \ldots, \langle x^t, b_t \rangle$ is monotone consistent, each $x^i$ and $z$ belong to $\{0, 1\}^m$, and $w^\phi \cdot \phi(z) \geq \theta$ if and only if $|F^{-1}(1)| \geq K$.

The Winnow variables $x_1, \ldots, x_m$ are divided into three sets $A, B$ and $C$ where $A = \{x_1, \ldots, x_n\}$, $B = \{x_{n+1}, \ldots, x_{n+U}\}$ and $C = \{x_{n+U+1}, \ldots, x_m\}$. The unlabeled example $z$ is $1^{n+U} 0^{m-n-U}$, i.e. all variables in $A$ and $B$ are set to 1 and all variables in $C$ are set to 0. We thus have $w^\phi \cdot \phi(z) = M_A + M_B + M_{AB}$ where $M_A = \sum_{\emptyset \neq T \subseteq A} w_T^\phi$, $M_B = \sum_{\emptyset \neq T \subseteq B} w_T^\phi$ and $M_{AB} = \sum_{T \subseteq A \cup B, T \cap A \neq \emptyset, T \cap B \neq \emptyset} w_T^\phi$. We refer to monomials $\emptyset \neq T \subseteq A$ as *type-A* monomials, monomials $\emptyset \neq T \subseteq B$ as *type-B* monomials, and monomials $T \subseteq A \cup B, T \cap A \neq \emptyset, T \cap B \neq \emptyset$ as *type-AB* monomials.

The example sequence $S$ is divided into four stages. Stage 1 results in $M_A \approx |F^{-1}(1)|$; as described below the $n$ variables in $A$ correspond to the $n$ variables in the CNF formula $F$. Stage 2 results in $M_A \approx \alpha^q |F^{-1}(1)|$ for some positive integer $q$ which we specify later. Stages 3 and 4 together result in $M_B + M_{AB} \approx \theta - \alpha^q K$. Thus the final value of $w^\phi \cdot \phi(z)$ is approximately $\theta + \alpha^q(|F^{-1}(1)| - K)$, so we have $w^\phi \cdot \phi(z) \geq \theta$ if and only if $|F^{-1}(1)| \geq K$.

Since all variables in $C$ are 0 in $z$, if $T$ includes a variable in $C$ then the value of $w_T^\phi$ does not affect $w^\phi \cdot \phi(z)$. The variables in $C$ are "slack variables" which (i) make Winnow perform the correct promotions/demotions and (ii) ensure that $S$ is monotone consistent.

**Stage 1: Setting $M_A \approx |F^{-1}(1)|$.** We define the following correspondence between truth assignments $y^T \in \{0, 1\}^n$ and monomials $T \subseteq A$ : $y_i^T = 0$ if and only if $x_i$ is not present in $T$. For each clause $y_{i_1} \vee y_{i_2}$ in $F$, Stage 1 contains $V$ negative examples such that $x_{i_1} = x_{i_2} = 0$ and $x_i = 1$ for all other $x_i \in A$. We show below that (1) Winnow makes a false positive prediction on each of these examples and (2) in Stage 1 Winnow never does a promotion on any example which has any variable in $A$ set to 1. Consider any $y^T$ such that $F(y^T) = 0$. Since our examples include an example $y^S$ such that $y^T \leq y^S$ the monomial $T$ is demoted at least $V$ times. As a result after Stage 1 we will have that for all $T$, $w_T^\phi = 1$ if $F(y^T) = 1$ and $0 < w_T^\phi \leq \alpha^{-V}$ if $F(y^T) = 0$. Thus we will have $M_A = |F^{-1}(1)| + \gamma_1$ for some $0 < \gamma_1 < 2^n \alpha^{-V} < \frac{1}{2}$.

We now show how the Stage 1 examples cause Winnow to make a false positive prediction on negative examples which have $x_{i_1} = x_{i_2} = 0$ and $x_i = 1$ for all other $i$ in $A$ as described above. For each such negative example in Stage 1 six new slack variables $x_{\beta+1}, \ldots, x_{\beta+6} \in C$ are used as follows: Stage 1 has $\lceil \log_\alpha(\theta/3) \rceil$ repeated instances of the positive example which has $x_{\beta+1} = x_{\beta+2} = 1$ and all other bits 0. These examples cause promotions which result in $\theta \leq w_{x_{\beta+1}}^\phi + w_{x_{\beta+2}}^\phi + w_{x_{\beta+1}x_{\beta+2}}^\phi < \alpha\theta$ and hence $w_{x_{\beta+1}}^\phi \geq \theta/3$. Two other groups of similar examples (the first with $x_{\beta+3} = x_{\beta+4} = 1$, the second with $x_{\beta+5} = x_{\beta+6} = 1$) cause $w_{x_{\beta+3}}^\phi \geq \theta/3$ and $w_{x_{\beta+5}}^\phi \geq \theta/3$. The next example in $S$ is the negative example which has $x_{i_1} = x_{i_2} = 0$, $x_i = 1$ for all other $x_i$ in $A$, $x_{\beta+1} = x_{\beta+3} = x_{\beta+5} = 1$ and all other bits 0. For this example $w^\phi \cdot \phi(x) > w_{x_{\beta+1}}^\phi + w_{x_{\beta+3}}^\phi + w_{x_{\beta+5}}^\phi \geq \theta$ so Winnow makes a false positive prediction.

Since $F$ has at most $n^2$ clauses and there are $V$ negative examples per clause, this construction can be carried out using $6Vn^2$ slack variables $x_{n+U+1}, \ldots, x_{n+U+6Vn^2}$. We thus have (1) and (2) as claimed above.

**Stage 2: Setting $M_A \approx \alpha^q |F^{-1}(1)|$.** The first Stage 2 example is a positive example with $x_i = 1$ for all $x_i \in A$, $x_{n+U+6Vn^2+1} = 1$ and all other bits 0. Since each of the $2^n$ monomials which contain $x_{n+U+6Vn^2+1}$ and are satisfied by this example have $w_T^\phi = 1$, we have $w^\phi \cdot \phi(x) = 2^n + |F^{-1}(1)| + \gamma_1 < 2^{n+1}$. Since $\theta > 2^m/\text{poly}(m) > 2^{n+1}$ (recall from equation (2) that $m > 6n^2$), after the resulting promotion we have $w^\phi \cdot \phi(x) = \alpha(2^n + |F^{-1}(1)| + \gamma_1) < \alpha 2^{n+1}$. Let

$$q = \lceil \log_\alpha(\theta/2^{n+1}) \rceil - 1$$

so that

$$\alpha^q 2^{n+1} < \theta \leq \alpha^{q+1} 2^{n+1}. \tag{3}$$

Stage 2 consists of $q$ repeated instances of the positive example described above. After these promotions we have $w^\phi \cdot \phi(x) = \alpha^q(2^n + |F^{-1}(1)| + \gamma_1) < \alpha^q 2^{n+1} < \theta$. Since $1 < |F^{-1}(1)| + \gamma_1 < 2^n$ we also have

$$\alpha^q < M_A = \alpha^q(|F^{-1}(1)| + \gamma_1) < \alpha^q 2^n < \theta/2. \tag{4}$$

Equation (4) gives the value which $M_A$ will have throughout the rest of the argument.

**Some Calculations for Stages 3 and 4.** At the start of Stage 3 each type-$B$ and type-$AB$ monomial $T$ has $w_T^\phi = 1$. There are $n$ variables in $A$ and $U$ variables in $B$ so at the start of Stage 3 we have $M_B = 2^U - 1$ and $M_{AB} = (2^n - 1)(2^U - 1)$. Since no example in Stages 3 or 4 satisfies any $x_i$ in $A$, at the end of Stage 4 $M_A$ will still be $\alpha^q(|F^{-1}(1)| + \gamma_1)$ and $M_{AB}$ will still be $(2^n - 1)(2^U - 1)$. Therefore at the end of Stage 4 we have

$$w^\phi \cdot \phi(z) = M_B + \alpha^q(|F^{-1}(1)| + \gamma_1) + (2^n - 1)(2^U - 1).$$

To simplify notation let

$$D = \theta - (2^n - 1)(2^U - 1) - \alpha^q K.$$

Ideally at the end of Stage 4 the value of $M_B$ would be $D - \alpha^q \gamma_1$ since this would imply that $w^\phi \cdot \phi(z) = \theta + \alpha^q(|F^{-1}(1)| - K)$ which is at least $\theta$ if and only if $|F^{-1}(1)| \geq K$. However it is not necessary for $M_B$ to assume this exact value, since $|F^{-1}(1)|$ must be an integer and $0 < \gamma_1 < \frac{1}{2}$. As long as

$$D \leq M_B < D + \frac{1}{2}\alpha^q \tag{5}$$

we get that

$$\theta + \alpha^q(|F^{-1}(1)| - K + \gamma_1) < w^\phi \cdot \phi(z) < \theta + \alpha^q(|F^{-1}(1)| - K + \gamma_1 + \frac{1}{2}).$$

Now if $|F^{-1}(1)| \geq K$ we clearly have $w^\phi \cdot \phi(z) \geq \theta$. On the other hand if $|F^{-1}(1)| < K$ then since $|F^{-1}(1)|$ is an integer value $|F^{-1}(1)| \leq K - 1$ and we get $w^\phi \cdot \phi(z) < \theta$. Therefore all that remains is to construct the examples in Stages 3 and 4 so that that $M_B$ satisfies Equation (5).

We next calculate an appropriate granularity for $D$. Note that $K \leq 2^n$, so by Equation (3) we have that $\theta - \alpha^q K > \theta/2$. Now recall from Equations (2) and (1) that $m > n + U + 6n^2$ and $\theta > 2^m/\text{poly}(m)$, so $\theta/2 \geq 2^{n+U+6n^2}/\text{poly}(m) \gg 2^n 2^U$. Consequently we certainly have that $D > \theta/4$, and from Equation (3) we have that $D > \theta/4 > \alpha^q 2^{n-1} > \frac{1}{4}\alpha^q$. Let

$$c = \lceil \log_\alpha 4 \rceil,$$

so that we have

$$\alpha^{q-c} \leq \frac{1}{4}\alpha^q < D. \tag{6}$$

There is a unique smallest positive integer $p > 1$ which satisfies $D \leq p\alpha^{q-c} < D + \frac{1}{4}\alpha^q$. The Stage 3 examples will result in $M_B$ satisfying $p < M_B < p + \frac{1}{4}$. We now have that:

$$\alpha^{q-c} < D \leq p\alpha^{q-c} \quad < \quad D + \frac{1}{4}\alpha^q$$
$$\leq \quad \theta - \frac{3}{4}\alpha^q \tag{7}$$
$$\leq \quad \alpha^{q+1}2^{n+1} - 3\alpha^{q-c} \tag{8}$$
$$= \quad \alpha^{q-c} \cdot (\alpha^{c+1}2^{n+1} - 3). \tag{9}$$

Here (7) holds since $K \geq 1$, and thus (by definition of $D$) we have $D + \alpha^q \leq \theta$ which is equivalent to Equation (7). Inequality (8) follows from Equations (6) and (3).

Hence we have that

$$1 < p \leq \alpha^{c+1}2^{n+1} - 3 \leq 2^{n+1+\lceil (c+1)\log \alpha \rceil} - 3 = 2^U - 3, \tag{10}$$

where the second inequality in the above chain follows from Equation (9). We now use the following lemma:

**Lemma 10** *For all $\ell \geq 1$, for all $1 \leq p \leq 2^{\ell} - 1$, there is a monotone CNF $F_{\ell,p}$ over $\ell$ Boolean variables which has at most $\ell$ clauses, has exactly $p$ satisfying assignments in $\{0, 1\}^{\ell}$, and can be constructed from $\ell$ and $p$ in $poly(\ell)$ time.*

**Proof:** The proof is by induction on $\ell$. For the base case $\ell = 1$ we have $p = 1$ and $F_{\ell,p} = x_1$. Assuming the lemma is true for $\ell = 1, \ldots, k$ we now prove it for $\ell = k + 1$ :

If $1 \leq p \leq 2^k - 1$ then the desired CNF is $F_{k+1,p} = x_{k+1} \wedge F_{k,p}$. Since $F_{k,p}$ has at most $k$ clauses $F_{k+1,p}$ has at most $k + 1$ clauses. If $2^k + 1 \leq p \leq 2^{k+1} - 1$ then the desired CNF is $F_{k+1,p} = x_{k+1} \vee F_{k,p-2^k}$. By distributing $x_k$ over each clause of $F_{k,p-2^k}$ we can write $F_{k+1,p}$ as a CNF with at most $k$ clauses. If $p = 2^k$ then $F_{k,p} = x_1$. $\qquad\square$

**Stage 3: Setting $M_B \approx p$.** Let $F_{U,p}$ be an $r$-clause monotone CNF formula over the $U$ variables in $B$ which has $p$ satisfying assignments. Similar to Stage 1, for each clause of $F_{U,p}$, Stage 3 has $W$ negative examples corresponding to that clause, and as in Stage 1 slack variables in $C$ are used to ensure that Winnow makes a false positive prediction on each such negative example. Thus the examples in Stage 3 cause $M_B = p + \gamma_2$ where $0 < \gamma_2 < 2^U \alpha^{-W} < \frac{1}{4}$. Since six slack variables in $C$ are used for each negative example and there are $rW \leq UW$ negative examples, the slack variables $x_{n+U+6Vn^2+2}, \ldots, x_{m-2}$ are sufficient for Stage 3.

**Stage 4: Setting $M_B + M_{AB} \approx \theta - \alpha^q K$.** All that remains is to perform $q - c$ promotions on examples which have each $x_i$ in $B$ set to 1. This will cause $M_B$ to equal $(p + \gamma_2)\alpha^{q-c}$. By the inequalities established above, this will give us

$$D \leq p\alpha^{q-c} < (p + \gamma_2)\alpha^{q-c} = M_B < D + \frac{1}{4}\alpha^q + \gamma_2\alpha^{q-c} < D + \frac{1}{2}\alpha^q$$

which is as desired.

In order to guarantee $q - c$ promotions we use two sequences of examples of length $q - \lceil\frac{U-n}{\log\alpha}\rceil$ and $\lceil\frac{U-n}{\log\alpha}\rceil - c$ respectively. We first show that these are positive numbers. It follows directly from the definitions $U = n + 1 + \lceil(\lceil\log_\alpha 4\rceil + 1)\log\alpha\rceil$ and $c = \lceil\log_\alpha 4\rceil$ that $\frac{U-n}{\log\alpha} \geq c$. Since $\theta > 2^{6n^2}$ (by definition of $m$ and Equation (1)) and $\alpha$ is bounded by a polynomial in $m$, we clearly have that $\log(\theta/2^{n+1}) > U - n + \log(\alpha)$. Now since $q = \lceil\log_\alpha(\theta/2^{n+1})\rceil - 1$ this implies that $q > \frac{\log(\theta/2^{n+1})}{\log(\alpha)} - 1 > \frac{U-n}{\log\alpha}$, so that $q - \lceil\frac{U-n}{\log\alpha}\rceil > 0$.

The first $q - \lceil\frac{U-n}{\log\alpha}\rceil$ examples in Stage 4 are all the same positive example which has each $x_i$ in $B$ set to 1 and $x_{m-1} = 1$. The first time this example is received, we have $w^\phi \cdot \phi(x) = 2^U + p + \gamma_2 < 2^{U+1}$. Since $\theta > 2^{6n^2}$, by inspection of $U$ we have $2^{U+1} < \theta$, so Winnow performs a promotion. Similarly, after $q - \lceil\frac{U-n}{\log\alpha}\rceil$ occurrences of this example, we have

$$w^\phi \cdot \phi(x) = \alpha^{q-\lceil\frac{U-n}{\log\alpha}\rceil}(2^U + p + \gamma_2) < \alpha^{q-\lceil\frac{U-n}{\log\alpha}\rceil}2^{U+1} \leq \alpha^q 2^{n+1} < \theta$$

so promotions are indeed performed at each occurrence, and

$$M_B = \alpha^{q-\lceil\frac{U-n}{\log\alpha}\rceil}(p + \gamma_2).$$

The remaining examples in Stage 4 are $\lceil\frac{U-n}{\log\alpha}\rceil - c$ repetitions of the positive example $x$ which has each $x_i$ in $B$ set to 1 and $x_m = 1$. If promotions occurred on each repetition of this example then we would have $w^\phi \cdot \phi(x) = \alpha^{\lceil \frac{U-n}{\log \alpha} \rceil - c}(2^U + \alpha^{q - \lceil \frac{U-n}{\log \alpha} \rceil}(p + \gamma_2))$, so we need only show that this quantity is less than $\theta$. We reexpress this quantity as $\alpha^{\lceil \frac{U-n}{\log \alpha} \rceil - c}2^U + \alpha^{q-c}(p + \gamma_2)$. We have

$$
\begin{aligned}
\alpha^{q-c}(p + \gamma_2) \quad &< \quad p\alpha^{q-c} + \frac{1}{4}\alpha^{q-c} \\
&\leq \quad \theta - \frac{3}{4}\alpha^q + \frac{1}{16}\alpha^q \\
&< \quad \theta - \frac{1}{2}\alpha^q
\end{aligned}
\tag{11}
$$

where (11) follows from (7) and the definition of $c$. Finally, we have that $\alpha^{\lceil \frac{U-n}{\log \alpha} \rceil - c}2^U \leq \alpha \cdot 2^{2U-n-c\log \alpha} < \alpha \cdot 2^{2U-n-2} < \frac{1}{2\alpha} \frac{\theta}{2^{n+1}} < \frac{1}{2}\alpha^q$, where the last inequality is by Equation (3) and the previous inequality is by inspection of the values of $\alpha$, $\theta$ and $U$. Combining the two bounds above we see that indeed $w^\phi \cdot \phi(x) < \theta$.

Finally, we observe that by construction the example sequence $S$ is monotone consistent. Since $m = \text{poly}(n)$ and $S$ contains $\text{poly}(n)$ examples the transformation from M2SAT to KWP($\alpha, \theta$) is polynomial-time computable and the theorem is proved. □(Theorem 9)

## 5. Conclusion

Linear threshold functions are a weak representation language for which we have interesting learning algorithms. Therefore, if linear learning algorithms are to learn expressive functions, it is necessary to expand the feature space over which they are applied. This work explores the tradeoff between computational efficiency and convergence when using expanded feature spaces that capture conjunctions of base features.

We have shown that while each iteration of the kernel Perceptron algorithm can be executed efficiently, the algorithm can provably require exponentially many updates even when learning a function as simple as $f(x) = x_1 x_2 \ldots x_n$. On the other hand, the kernel Winnow algorithm has a polynomial mistake bound for learning polynomial-size monotone DNF, but our results show that under a widely accepted computational hardness assumption it is impossible to efficiently simulate the execution of kernel Winnow. The latter also implies that there is no general construction that will run Winnow using kernel functions.

Our results indicate that additive and multiplicative update algorithms lie on opposite extremes of the tradeoff between computational efficiency and convergence; we believe that this fact could have significant practical implications. By demonstrating the provable limitations of using kernel functions which correspond to high-degree feature expansions, our results also lend theoretical justification to the common practice of using a small degree in similar feature expansions such as the well-known polynomial kernel.[2]

Since the publication of the initial conference version of this work (Khardon, Roth, & Servedio, 2002), several authors have explored closely related ideas. One can show that our construction for the negative results for Perceptron does not extend (either in the PAC or online setting) to related algorithms such as Support Vector Machines which work by constructing a maximum margin hypothesis consistent with the examples. The paper (Khardon & Servedio, 2003) gives an analysis of the PAC learning performance of maximum margin algorithms with the monotone monomials kernel, and derives several negative results thus giving further negative evidence for the monomial kernel. In the paper (Cumby & Roth, 2003) a kernel for expressions in description logic (generalizing the monomials kernel) is developed and successfully applied for natural language and molecular problems. Takimoto and Warmuth (2003) study the use of multiplicative update algorithms other than Winnow (such as weighted majority) and obtain some positive results by restricting the type of loss function used to be additive over base features. Chawla *et al.* (2004) have studied Monte Carlo estimation approaches to approximately simulate the Winnow algorithm's performance when run over a space of exponentially many features. The use of kernel methods for logic learning and developing alternative methods for feature expansion with multiplicative update algorithms remain interesting and challenging problems to be investigated.

---

2. Our Boolean kernels are different than standard polynomial kernels in that all the conjunctions are weighted equally, and also in that we allow negations.

## Acknowledgments

This work was partly done while Khardon was at the University of Edinburgh and partly while Servedio was at Harvard University. The authors gratefully acknowledge financial support for this work by EPSRC grant GR/N03167, NSF grant IIS-0099446 and a Research Semester Fellowship Award from Tufts University (Khardon), NSF grants ITR-IIS-00-85836, ITR-IIS-0085980 and IIS-9984168 (Roth), and NSF grant CCR-98-77049 and NSF Mathematical Sciences Postdoctoral Fellowship (Servedio).